\pdfoutput=1

\documentclass[11pt]{article}

\usepackage{acl}

\usepackage{times}
\usepackage{latexsym}
\usepackage{booktabs}
\usepackage{microtype}
\usepackage{amssymb}
\usepackage{pifont}
\usepackage{xcolor} 
\usepackage{graphicx}
\usepackage{placeins}
\usepackage{natbib}
\usepackage{tablefootnote}
\usepackage{longtable}
\usepackage{enumitem}
\usepackage{tikz}
\usepackage{booktabs}
\usepackage{pdfpages}

\usepackage[T1]{fontenc}

\usepackage{tabularx}
\usepackage{multirow}
\usepackage{xspace}
\usepackage{subfigure}
\usepackage{caption}
\usepackage{url}
\usepackage{amsmath}
\usepackage{amssymb}
\usepackage{amsfonts}
\usepackage{comment}
\usepackage{array}
\usepackage{multirow}
\usepackage{microtype}
\usepackage{pgfplots}
\usepackage{soul}
\usepackage{xcolor}
\usepackage{enumitem}
\usepackage{tikz}
\usepackage{cleveref}
\usepackage{pgf-pie}  

\usepackage[many]{tcolorbox}        
\usepackage{lipsum}

\tcbset{
    sharp corners,
    colback = white,
    before skip = 0.2cm,    
    after skip = 0.5cm      
}                           

\newtcolorbox{boxA}{
    boxrule = 0.5pt,
    fontupper=\small,
    colframe = black 
}

\usepackage[textsize=footnotesize]{todonotes}
\usepackage{linguex}
\usepackage{subcaption}
\alignSubExtrue

\usepackage[verbose]{newunicodechar}

\usepackage{inconsolata}

\title{Building Better: Avoiding Pitfalls in Developing Language Resources\\ 
when Data is Scarce}

\author{Nedjma Ousidhoum$^1$, Meriem Beloucif$^2$, Saif M. Mohammad$^3$\\
$^1$Cardiff University, $^2$ Uppsala University, $^3$ National Research Council Canada \\
\texttt{OusidhoumN@cardiff.ac.uk} \texttt{ meriem.beloucif@lingfil.uu.se}\\ \texttt{ saif.mohammad@nrc-cnrc.gc.ca}
}
 
\begin{document}
 \maketitle

\begin{abstract}

Language is a form of symbolic capital that affects people's lives in many ways \cite{bourdieu1977economics,bourdieu1991language}. As a powerful means of communication, it reflects identities, cultures, traditions, and societies more broadly.
Therefore, data in a given language should be regarded as more than just a collection of tokens. Rigorous data collection and labeling practices are essential for developing more human-centered and socially aware technologies.
Although there has been growing interest in under-resourced languages within the NLP community, work in this area faces unique challenges, such as data scarcity and limited access to qualified annotators.

In this paper, we collect feedback from individuals directly involved in and impacted by NLP artefacts for medium- and low-resource languages. We conduct both quantitative and qualitative analyses of their responses and highlight key issues related to:
(1)\ data quality, including linguistic and cultural appropriateness; and
(2)\ the ethics of common annotation practices, such as the misuse of participatory research.
Based on these findings, we make several recommendations for creating high-quality language artefacts that reflect the cultural milieu of their speakers, while also respecting the dignity and labor of data workers.
\end{abstract}

\section{Introduction}

 \begin{figure}
    \centering
    \includegraphics[trim={5cm 0 5cm 0},clip,width=0.95\linewidth]{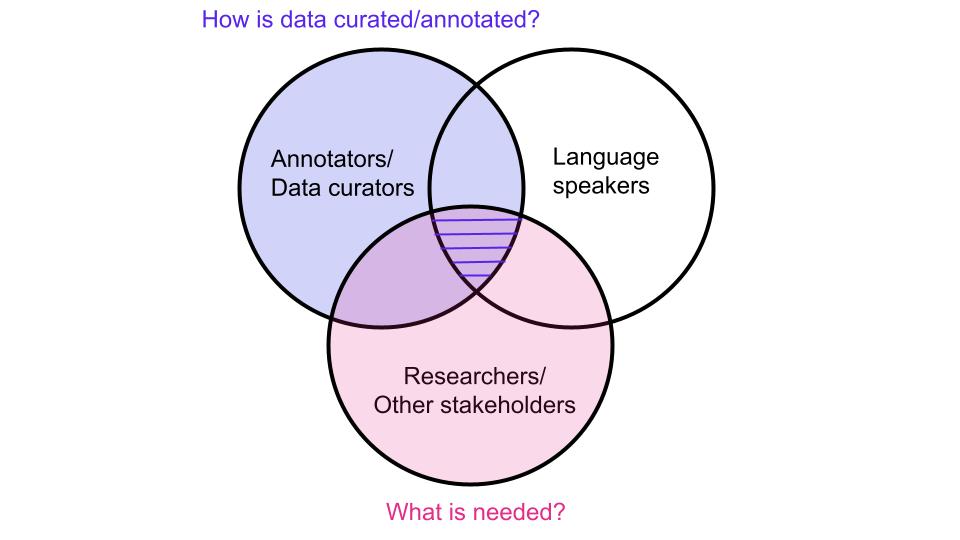}
    \caption{\textbf{The main stakeholders and their roles in a data annotation project:}
We conducted a survey and reached out to the CL community, specifically NLP researchers and practitioners who have worked on medium- to low-resource languages. Some of the questions focus on the perspectives of a subset of stakeholders highlighted in the figure---namely, speakers who focus on their own languages.
    }
    \label{fig:survey_respondents}
\end{figure}

Over the past few years, there has been growing interest in making NLP research more human-centered \cite{kotnis2022human-nlp} and socially aware \cite{yang2024call_socially_aware_tech}. As language technologies are deeply dependent on data quality \cite{Hirschberg-manning-article2015} and its alignment with the needs of speakers, researchers, and other users, incorporating diverse stakeholder perspectives is essential for building high-quality tools and resources.
Traditionally, decisions around data selection, collection, annotation, and model design have been made primarily by researchers. However, the involvement of native speakers, whose languages are at the core of these artefacts, is paramount to better design practices  \cite{bird2024centering}, since language is inseparable from culture and identity \cite{bourdieu1991language}.
Despite this, when it comes to mid- to low-resource languages, researchers often resort to using any available datasets, seldom scrutinising their quality or relevance due to resource scarcity. While NLP for English and other high-resource languages has benefited from improved standards for corpus quality and ethical research practices \cite{gebru2021datasheets,bender-tacl_a_00041,mohammad2022ethics}, these standards are not consistently extended to under-resourced languages \cite{joshi-etal-2020-state}.
Therefore, NLP artefacts for underrepresented languages often lack cultural and linguistic grounding, leading to false generalisations and systemic shortcomings \cite{bender-tacl_a_00041}. This disconnect may hinder meaningful progress and perpetuate inequalities \cite{blasi-etal-2022-systematic,held2023_nlp_coloniality_material}, produce sub-optimal user experiences, and reinforce long-standing language hierarchies \cite{kahane1986typology}.

In this position paper, we highlight the current limitations of NLP research for mid- to low-resource languages, specifically in terms of data collection, ethical annotation practices, and overall data quality. We reached out to the CL community involved in NLP projects on under-served languages and conducted a survey to report on the common incentives, limitations, applied norms, and practices (see Figure \ref{fig:survey_respondents}). 
We present the survey's results and provide a set of recommendations based on the responses, focusing on (1)\ fairness and centering of the language speakers, (2)\ choosing suitable data sources, (3)\ setting fair and realistic expectations when recruiting annotators, and (4)\ avoiding cultural misrepresentation.

\section{Related Work}
Work on ethical practices in AI, ML, and NLP research spans a wide range of topics, including artefact documentation \cite{bender2011achieving,bender-tacl_a_00041,gebru2021datasheets,rogers-etal-2021-just-think,mohammad2022ethics} and best practice recommendations \cite{hollenstein-etal-2020-towards,mohammad2023best}. 
Research specifically focused on low-resource languages tends to address the general state of NLP in this area \cite{held2023_nlp_coloniality_material,joshi-etal-2020-state,blasi-etal-2022-systematic,dogruoz-sitaram-2022-language}, data collection challenges \cite{yu-etal-2022-beyond}, limitations in specific tasks such as machine translation \cite{mager2023ethical}, developments in LLM research \cite{mihalcea2024ai}, and the fundamental issue of including the people whose languages are being studied \cite{mager2023ethical,bird-2020-decolonising,bird-2022-local,bird2024centering,lent-etal-2022-creole}.
Such work highlights the peculiarities of many low-resource languages, the majority of which are vernacular rather than institutionalised or written \cite{bird2024centering,bird2024must}. It further advocates for language communities to assume agency over their own languages \cite{schwartz-2022-primum,markl-etal-2024-language,mihalcea2024ai}. For instance, \citet{bird2024centering} examine how experts, such as linguists and computer scientists, collaborate with language communities through participatory design approaches \cite{winschiers2010being}, while \citet{cooper2024s} offer guidance on engaging with Indigenous communities beyond concerns of mere accuracy. \citet{dogruoz-sitaram-2022-language} shed light on the need to avoid treating NLP for low-resource languages as a scaled-down version of high-resource language technologies, emphasising the importance of accounting for linguistic and cultural peculiarities. Similarly, \citet{adebara-abdul-mageed-2022-towards} stress the significance of such language-specific features (e.g., tones) with a focus on African languages.
Beyond language speakers, other work considers the needs of users more broadly. For instance, \citet{blaschke2024dialect} address the concerns of dialect speakers and emphasise the importance of involving them in the development of language tools and resources. In addition, \citet{yang2024call_socially_aware_tech} define the concept of social awareness and advocate against treating language purely as a computational problem in NLP.

In this paper, we contribute to this ongoing discussion by shifting focus to the practical challenges faced by NLP researchers and practitioners working on mid- and low-resource languages by drawing on methods from social sciences \cite{cetina1999epistemic}. We investigate the methodological practices and issues currently shaping the field.
To the best of our knowledge, little research has examined how NLP work on low-resource languages engages with online communities, apart from a few case studies involving participatory frameworks (e.g., Masakhane) \cite{birhane2022power}, and the work of \citet{lent-etal-2022-creole}, who analyse 38 responses from Creole speakers about their experiences with language technologies. Based on an analysis of feedback from our survey respondents, we offer practical recommendations that prioritise transparency and ethically grounded approaches to building more human-centered NLP artefacts for under-served languages.

\section{Survey}
\begin{table*}[ht]
    \centering
    \begin{tabular}{cc|cc|cc}
    \toprule
        \textbf{Projects in}&\textbf{}&\textbf{Task}&\textbf{}&\textbf{Motivation}&\textbf{}\\
        \midrule
        Industry & 12\%&Data creation&47\%&Scientific interest& 81\%\\
        Academia & 57\% &Data annotation&33\%&Building language technologies&72\%\\
        Both &31\%&Data collection&33\%&Limitations in language(s) of interest&60\%\\
        &&Model construction&9\%&LLM research&59\%\\
       \bottomrule
    \end{tabular}
    \caption{Reported project affiliations, tasks in which the annotators were involved, and the different motivations or incentives. Note that percentages do not sum up to one as respondents could report on more than one project in both industry and academia.}
    \label{tab:industry+academia}
\end{table*}
Our main goal was to investigate the current challenges and problematic practices in NLP research for mid- to low-resource languages and to propose potential solutions. To this end, we reached out to the NLP community (i.e., *CL networks) between June and October 2024 via platforms such as X (formerly Twitter), LinkedIn, Google Groups, Slack channels of online NLP communities, and direct emails. We specifically targeted researchers and annotators working on mid- and low-resource languages, language variants, dialects, and vernaculars, to survey how research in these areas is conducted. Participants reported on common practices, motivations, and key issues. We then conducted both quantitative and qualitative analyses of the responses.
\subsection{Respondents}
The respondents are NLP researchers and practitioners involved in data collection, annotation, model development, or other research questions related to under-served languages.
\subsection{Survey Structure}
We ask the respondents about (1)\ their previous experiences in the area, (2)\ current problems and limitations related to their language(s) of interest, (3)\ the motivation behind their involvement in various projects, and (4)\ how they were credited for tasks often specific to low-resource language research---such as annotation work conducted via online community forums or participatory frameworks.

Note that we allowed respondents to determine for themselves what constitutes a low-resource language, as there is no universally accepted definition. For example, most researchers would agree that Tamasheq is a low-resource language, whereas opinions may differ regarding Malaysian.

\subsubsection{General Questions}
Respondents had the option to provide their names and contact information for potential follow-up. They were asked about:

\begin{itemize}[noitemsep,nolistsep]
\item the language(s) they work on,
\item the project(s) they have been involved in,
\item whether they are or were part of any online community (i.e., participatory research framework),
\item whether the project(s) they worked on were based in industry, academia, or both,
\item the kinds of NLP tools that are or would be relevant and useful for their language(s) of interest,
\item their reasons for working on this/these language(s).
\end{itemize}

\subsubsection{Reporting on Incentives and Potential Limitations}
We looked into the common reasons why researchers work on low-resource languages. Therefore, we asked the participants to report on: 
\begin{itemize}[noitemsep,nolistsep]
    \item the incentive(s) for working on their language(s) of interest, 
    \item the incentive(s) for working on specific project(s) or task(s).
\end{itemize}

As we are aware of potential drawbacks in NLP for mid- to low-resource languages \cite{blasi-etal-2022-systematic}, we examined whether the respondents had been working in the area due to any limitations observed in available NLP tools in their language(s) of interest. Note that these questions were optional as researchers may work on any language for various other reasons. 
We asked the participants to report on:

\begin{itemize}[noitemsep,nolistsep]
    \item any observed limitations and optionally list some tools or resources in their language(s) of interest as examples, 
    \item potential language-specific challenges in their language(s) of interest.
    
\end{itemize}

\subsubsection{Reporting on Credit Attribution}
We asked the respondents about how often they were properly credited for their work. Moreover, since reaching out to online participatory frameworks is common to projects that focus on under-resourced languages, we asked whether the participants were involved in past projects through such frameworks. This is because involving communities in NLP and ML projects is relatively new to the field and can therefore be abused (purposefully or not) due to the lack of clear standards regarding data workers in such contexts. 
Therefore, our questions were the following:

\begin{itemize}[noitemsep,nolistsep]
    \item How often did the respondents receive credit for their contributions? E.g., financial compensation for annotating data. 
    \item How often were they offered authorship when making substantial contributions to the data collection and/or data annotation?
    \item What were their incentives for projects in which they did not receive financial compensation or authorship?
    \item How long did the process take especially when they were not properly credited?
\end{itemize}

\section{Findings}\label{sec:findings}

We received 81 responses from researchers working on a wide range of languages and language families. 
Even though including contact information was optional, more than 90\% of the respondents chose not to reply anonymously, and 80\% asked for updates on the project. 
Table \ref{tab:industry+academia} shows the distribution of responses to questions on project affiliations, the tasks in which the respondents were involved, and their motivations for working on under-resourced languages. Note that percentages do not sum up to 100\% as respondents could report on more than one project. That is, participants could also be involved in several languages and NLP tasks.
As shown in Table \ref{tab:industry+academia}, most participants were involved in dataset curation mainly motivated by scientific interest or curiosity, and for building language technologies because of observed limitations in resources dedicated to their language(s) of interest.

\begin{figure}
    \centering
    \includegraphics[trim={0cm 0 0cm 0},clip,width=0.99\linewidth]{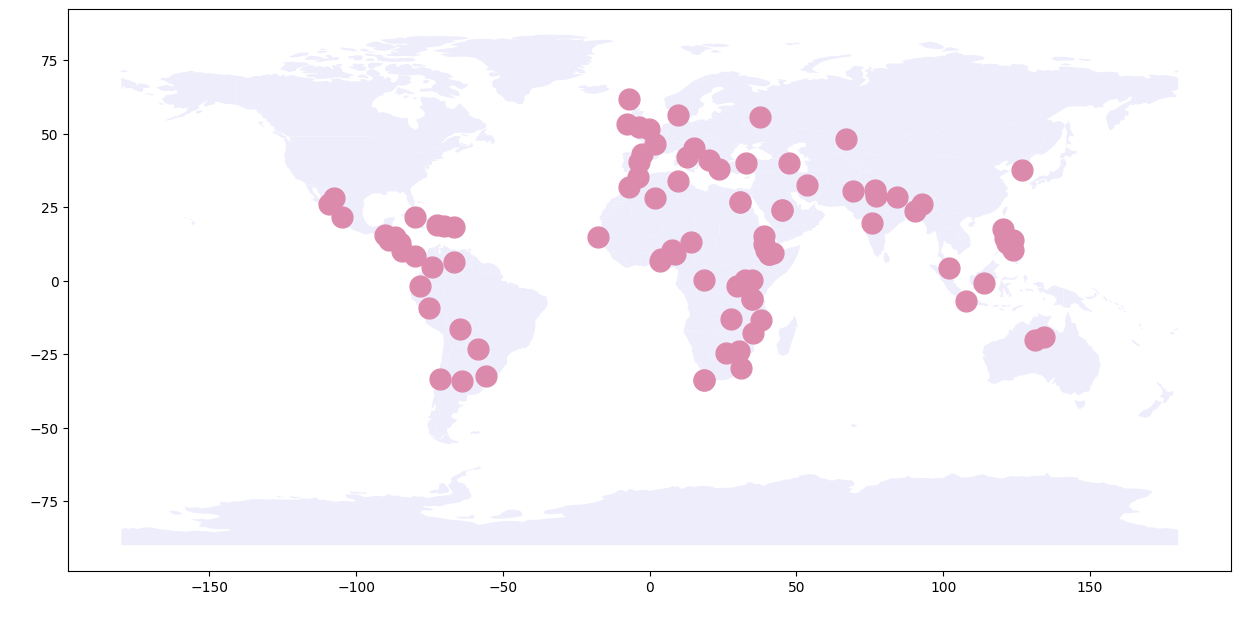}
    \caption{\textbf{The main locations} where the languages of interest of our respondents are spoken.}
    \label{fig:survey_respondents_locations}
\end{figure}

\subsection{General Information}
\subsubsection{Projects}
Respondents could report on one or multiple projects they had been involved in. As shown in Table~\ref{tab:industry+academia}, most respondents reported working on academic projects. Around one-third participated in collaborations between industry and academia, or were involved in both types of projects.\footnote{Note that although over 50\% of respondents named the projects they participated in, we do not disclose these in order to protect their anonymity.}

\subsubsection{Languages}
Among the 81 responses, respondents reported working on over 70 low-resource languages, which they specifically named (see Appendix). Figure~\ref{fig:survey_respondents_locations} illustrates the main regions where these languages are spoken.
These include variants, dialects, and vernaculars (e.g., country-specific Arabic dialects), mid- to low-resource languages (e.g., Indonesian), as well as widely acknowledged low-resource languages such as Welsh, Yoreme Nokki, and Setswana.
Additionally, around 12\% of respondents reported working on language families or branches, such as South Asian languages, all Gaeilige dialects, or Arabic/English varieties.
A significant proportion of respondents also work on high-resource languages in parallel.

\subsection{Incentives and Potential Limitations}
When asking respondents why they had worked on NLP for under-resourced languages, we provided a checklist from which they could select multiple options and add their own entry.
We report on common motivations and practices that are typically specific to mid- to low-resource settings, often due to factors such as data scarcity. We also identify problematic instances and analyse the potential reasons behind some of them.

\begin{figure}[t]
    \centering
    \includegraphics[trim={0cm 0cm 0cm 0cm},clip,width=0.99\linewidth]{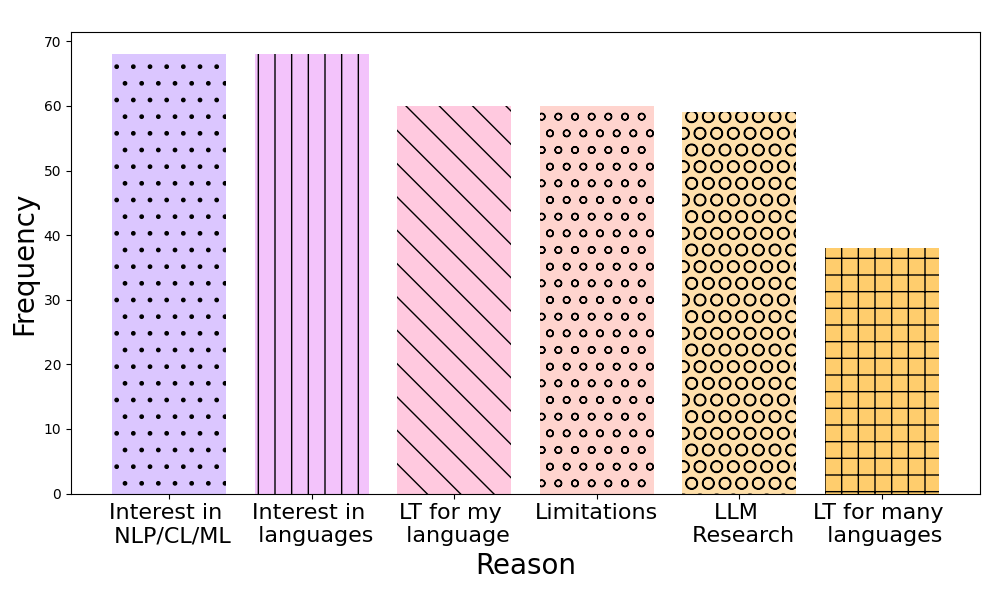}
    \caption{\textbf{Frequency of each incentive}. Note that the percentages do not sum up to 100 as the respondents could choose more than one option.}
    \label{figure:incentives_found}
\end{figure}
When further examining the common motivations, we report more detailed numbers in Figure \ref{figure:incentives_found}. Among those who were motivated by scientific curiosity or interest in Table \ref{tab:industry+academia} there were those whose interest was in NLP/CL/ML research (68\%) and those whose interest was in languages (68\%). Note that the two are not mutually exclusive.
For the respondents whose motivation was to build language technologies, most of them were more interested in building technologies for their own language(s) (60\%) as opposed to building technologies for as many languages as possible (38\%). This is particularly interesting as it constitutes evidence of the power of language as a symbolic capital \cite{bourdieu1991language}, which can sometimes manifest in the feeling of ``a duty'' that one has towards their language.

Other frequent motivations include marked limitations in language resources and tools in the language(s) of interest (60\%) and the willingness to contribute to LLM research (59\%).
\subsubsection{Reported Limitations}\label{subsubsec:marked_limitations}
\begin{figure}[t]
    \centering
    \includegraphics[trim={0cm 0cm 0cm 0cm},clip,width=0.99\linewidth]{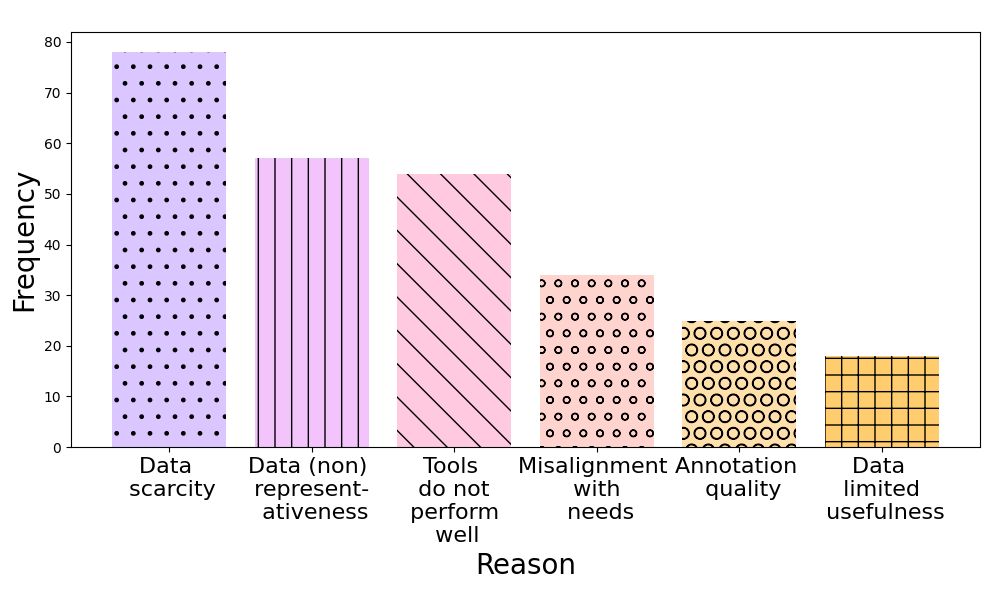}
    \caption{\textbf{Frequency of each reported limitation} when the respondents reported working on NLP for low-resource languages due to marked shortcomings.}
    \label{figure:limitations_found}
\end{figure}

More than 60\% of the respondents reported working on low-resource languages due to marked limitations in currently available resources for their language(s) of interest. 
To shed light on these limitations, we showed the respondents a predefined list of possible shortcomings 
as well as a text box where they could add any observed limitations.
As shown in Figure \ref{figure:limitations_found}: the predominant limitation is data scarcity (78\%); followed by the lack of representativeness of the data (58\%) that can manifest in, e.g., unnatural or translationese data instances; the under-performance of the available tools (54\%); their misalignment with the users' needs (54\%); the low quality of the annotations (25\%); and the lack of the usefulness of the data (18\%).

\subsubsection{Qualitative Analysis of the Limitations}
\begin{figure}[!h]
    \centering
    \scalebox{0.72}{
    \includegraphics[trim={0.5cm 2cm 0.5cm 2cm},clip,width=0.99\linewidth]{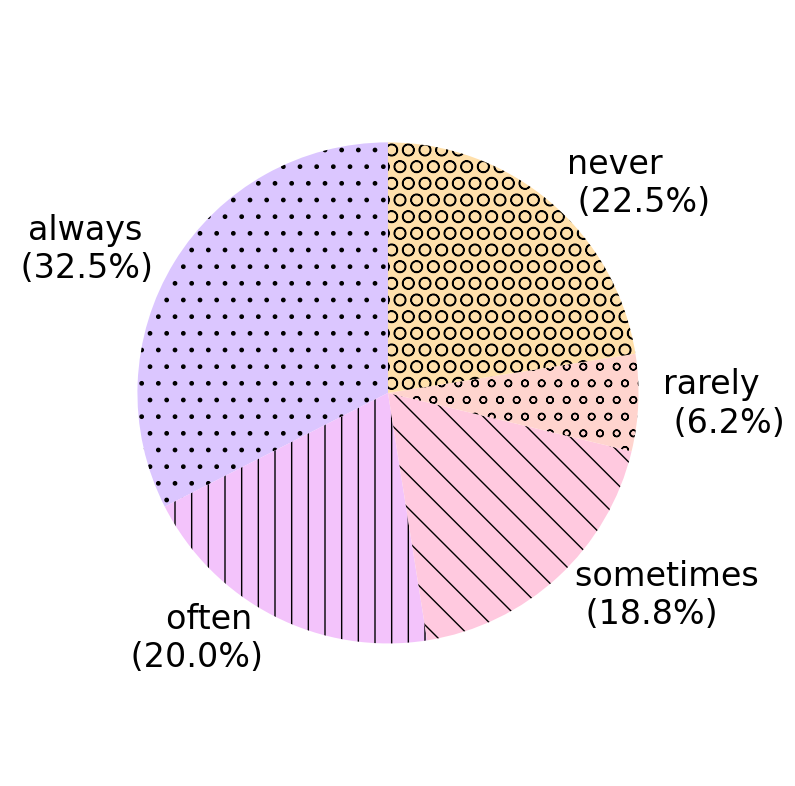}
    }
    \caption{\textbf{Respondents on getting credit} for projects they were involved in.}
    \label{fig:survey_respondents_credit}
\end{figure}
We provided the respondents with free text sections where they could report examples of tools or resources that suffer from the limitations they mentioned to justify their choices. When manually processing the answers, we noticed the following recurring topics:
\begin{enumerate}[noitemsep,nolistsep]
\item \textbf{Limitations related to existing resources:} such as their public unavailability, small size, or limited representativeness and quality.
\item \textbf{Limitations related to the practices adopted when building new resources:} such as:
\begin{itemize}
    \item the reliance on machine translation tools and LLMs to build resources;
    \item the lack of awareness of culture-specific and linguistic challenges of the languages in question;
    \item the challenges with annotator recruitment due to the lack of availability of native or near-native speakers on commonly used annotation platforms (e.g., AMT and Prolific), 
    \item the potential misuse of online communities and participatory framework services.
\end{itemize}
\item \textbf{Fundamental problems related to NLP research on mid- to low-resource languages:} such as the lack of funding often due to the ``low prestige'' language dilemma---the false notion that some languages or language varieties are more important than others. Interestingly, this was equally observed in projects from academia and industry.
\end{enumerate}
\noindent We discuss all three of these themes below.

\paragraph{Currently Available Resources}
Since many under-resourced languages are not institutional but rather vernacular \cite{bird2024centering}, collecting data presents considerable challenges when one solely relies on textual data for, e.g., Bantu languages.
Further, the focus on English and the reliance on translated data harms the quality of the generated datasets as they do not capture the subtle peculiarities of a given language.
It is important to note that what is translated and whether it was further verified by a fluent speaker makes a difference as translating Wikipedia texts can be easier than translating conversational, informal, or religious texts \cite{hutchinson-2024-modeling}.

\paragraph{Limitations with respect to Building New Resources}
Lack of representativeness and unnaturalness of the data were commonly reported in the responses. The respondents reported a lack of awareness of language variants and cultural aspects when building a language-specific artefact; the reliance on the standardised version of a given language due to power dynamics (i.e., more power in the hands of well-funded institutions and established researchers); the presence of offensive utterances in the data due to a lack of data filtering; and potentially wrong assumptions about a language or a culture. 
For instance, the time-specific context and usage of some languages, such as ancestral ones (e.g., Coptic), have considerably changed and one has to take these facts into account.
In addition, datasets may be collected from inadequate sources or could be aligned with Western values, standards, or expectations. This can be due to power differentials or a lack of deeper examination carried along with locals and native speakers. Finally, researchers rely on personal connections as it is hard to impossible to find fluent speakers of mid- to low-resource languages on commonly used annotation platforms such as Amazon Mechanical Turk and Polific. Added to this reason, the lack of funding leads researchers to turn to participatory frameworks. This practice has been at the center of major NLP contributions in recent years \cite{birhane2022power}. However, despite its benefits for people with common research interests, the absence of well-established standards puts vulnerable community members at risk as their efforts may not be properly recognised.

\paragraph{Fundamental Problems}
Many respondents reported that conducting research in mid- to low-resource languages often entails high costs of data curation and potential outreach to local communities. Further, when resources for an under-served language exist, they are often not freely available.

\subsection{Credit Attribution}

\begin{figure}
    \centering
    \includegraphics[trim={0cm 0cm 0cm 0cm},clip,width=0.99\linewidth]{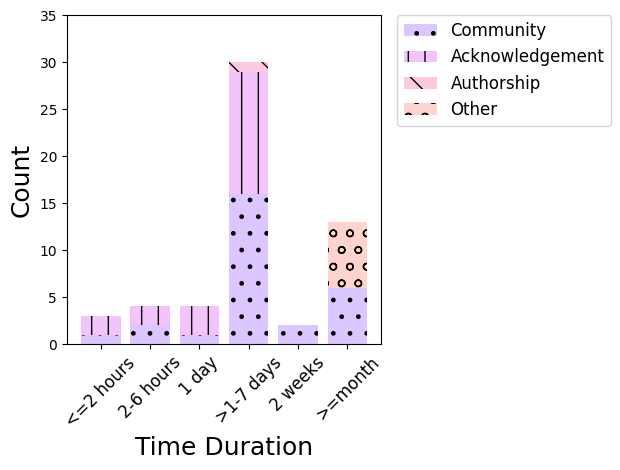}
    \caption{\textbf{Respondents on incentives when no proper credit (e.g., financial compensation for data annotation) was offered.} We show the counts of the incentives and the time it took participants to complete their work for a given project (from <=2 hours to more than a month).}
    \label{fig:problematic_incentives}
\end{figure}

We asked the respondents to share whether they were properly credited for their work by, for instance, getting financial compensation for a long annotation task, getting involved in the writing of a research paper for a resource that they built, etc. As shown in Figure \ref{fig:survey_respondents_credit}, most respondents (>67\%) report this not being the case at least once. Figure \ref{fig:problematic_incentives} shows the distributions of responses pertaining to how the respondents were incentivised to perform an annotation task for which they were eventually not given due credit.

\paragraph{Problematic Incentivisation}

For the respondents who reported that they did not receive proper credit for projects in which they were involved, we list the main incentives for joining these projects and the time it took the participants to complete the work. As shown in Figure \ref{fig:problematic_incentives}, they were either:
\begin{enumerate}[noitemsep,nolistsep]
    \item a member of a community or a participatory framework (see paragraph below),
    \item acknowledged on the website or the research paper, or
    \item somehow manipulated into thinking that there was a professional benefit in joining without proper compensation as we explain below.
\end{enumerate}

\paragraph{The Issue with Over-Reliance on Online Communities and Participatory Research}
When using standard crowdsourcing platforms such as AMT or Prolific, researchers can operationalise the annotation process for a given task. Despite their shortcomings \cite{fort2011amazon,irani2015cultural}, it is possible to protect workers by implementing tests and training for complex annotation tasks.
However, for mid- to low-resource languages, platforms such as AMT and Prolific often lack sufficient numbers of registered speakers. As a result, researchers frequently resort to personal networks or participatory approaches.
On the one hand, personal connections and community engagement can foster a sense of inclusion and help build trust. On the other hand, however, these approaches carry a significant risk of exploitation and emotional manipulation. That is, some junior researchers were clearly exploited, having been previously told that joining an online community for building language resources was prestigious and worth adding to their CVs. Some respondents expressed frustration at being offered, for instance, company merchandise in exchange for months of labor---a situation that raises further concerns about the involvement of big tech in NLP research for low-resource languages \cite{abdalla-etal-2023-elephant}.
As shown in Figure \ref{fig:problematic_incentives}, 40\% of respondents who spent between one day and over a month annotating data reported negative experiences. Their contributions were not adequately compensated, acknowledged, or recognised. This highlights the urgent need for clear guidelines and standards when engaging in community-based or participatory research efforts.
\section{Recommendations}
While there has been a considerable amount of work on the ethics of best practices for building NLP and ML artefacts \cite{bender-tacl_a_00041,mohammad2022ethics,mohammad2023best,leech2024questionable}, our findings substantiate the fact that research on under-resourced languages presents additional challenges linked to the reliance on unconventional practices.

\subsection{Center the People}
Our findings show that various issues ought to be addressed early as research in the area lacks established standards and is subject to power differentials. For example, some researchers reported that their work and contributions were diminished by more prominent individuals taking first authorship on papers to which they had contributed little or not at all. Many under-resourced languages are from what is called ``the Global South'' with a large number of them being spoken rather than written.

\paragraph{Speakers}\label{rec:speakers}

Language is an important part of a population's identity and technologies dealing with it have a direct impact on people's lives. Past NLP work highlights how to engage with speakers and communities whose languages are in question \cite{bird-2020-decolonising,bird2024centering, bird2024must,cooper2024s,ramponi-2024-language}.
We further reinforce this argument with our findings, i.e., most respondents report marked limitations in their languages (see Figure \ref{figure:limitations_found}). 
Hence, when a researcher reaches out to a group of people with little background knowledge of their culture or language, one needs to approach these problems from the perspective of the community in question \cite{bird-2022-local}. The question of \textbf{who is exactly served} needs to be addressed early to avoid any misconception of perceived needs for language technologies. 

\paragraph{Researchers vs. Data Workers}
In addition to the large proportion of survey respondents who reported not being properly credited for their labor (Figure~\ref{fig:survey_respondents_credit}), there were instances of emotional manipulation. This included appeals suggesting that one's labor would benefit the speakers of the language, and that this alone should be considered sufficient compensation. Notably, such practices were common in participatory research projects initiated by both small groups and Big Tech companies.
One has to set rules and expectations with clear communication on the purpose of a given research project. For instance, when dealing with online participatory research communities for data collection and annotation, extra care needs to be shown and benevolent prejudice such as depicting oneself as a savior of a local community \cite{bird-2022-local} must be avoided. 

The question of \textbf{who is annotating what} has to be addressed as well. The scarcity of qualified annotators can result in poor decision-making. Native speakers are often difficult to find online, which has led some researchers to select individuals from broadly associated regions---people who do not necessarily speak the specific language variant in question. This results in problematic overgeneralisations and overly simplistic solutions, where distinct languages or dialects are grouped together simply because they share one or a few attributes. For example, while Arabic dialects vary significantly, numerous research projects have treated entire regions, such as North Africa, as a linguistic monolith, sometimes to appear to have more data than is truly available.

\subsection{Be Fair: Give Credit where Credit is Due}
Our findings highlight an unfortunate trend---data workers and NLP researchers, especially those collaborating on participatory research projects, often suffer from a lack of recognition (Figure \ref{figure:incentives_found}). Accordingly, our recommendation is to set fair and comprehensive practices in participatory research projects, while considering power differentials.

\paragraph{Monetary Compensation}
Annotators must be provided with proper financial compensation. Companies and research labs that rely on communities for annotation and data creation should ensure fair compensation for contributors, for example, through legally binding contracts. Ideally, resource papers should include evidence that annotators were paid and treated fairly, as recommended by \citet{rogers-etal-2021-just-think}.

\paragraph{Co-authorship Standards}
Existing authorship standards must be followed and discussed before the start of a project, particularly with respect to whether data workers should be listed as authors. This is especially important for junior researchers who contribute significantly to resource construction and may expect a leading role in the resulting publication.
To inclusively recognise the contributions of junior researchers and those involved in data curation, we recommend the following:
\begin{enumerate}[noitemsep,nolistsep]
\item Credit all contributors involved in dataset creation, including those that contributed to data curation, design of the annotation setup, and the writing of the paper---as authors.
\item Let data annotators know that they are welcome to play a greater role in the project, possibly rising to the level of co-authors. Discuss with them that meaningful contributions to one or more of the following can lead to co-authorship:
\begin{itemize}
\item Participating in data curation and experimental design.
\item Running language-specific experiments and performing (language-specific) error analyses.
\item Conducting (language-specific) ablation studies.
\item Contributing to the writing of specific paper sections, such as the related work.
\item Providing insights into the resource by:
\begin{itemize}
\item Selecting culturally relevant or representative examples.
\item Offering explanations and interpretations.
\item Describing the annotation process and sharing key observations or challenges.
\end{itemize}
\end{itemize}
\end{enumerate}
\noindent These tips are adapted from large past annotations efforts that we led \cite{ousidhoum2024semrel2024,muhammad2025brighterbridginggaphumanannotated}.

\subsection{Choose the Jargon Carefully and Be Aware of False Generalisations}
As previously discussed in \ref{rec:speakers}, it is important to embrace social awareness and avoid grouping people from colonial and Western perspectives \cite{bird-2020-decolonising,bird-2022-local,held2023_nlp_coloniality_material}. 
Hence, one can avoid dismissive and outdated terms and classifications, e.g., ``the rest of the world''. Note also that \textit{The World's Values Survey} classification \cite{haerpfer2012world}, which is often used in NLP papers (e.g., \citet{santy-etal-2023-nlpositionality}), presents an orientalist view of the world \cite{said1978orientalism}. It has clear flaws such as including Christian-majority countries (e.g., Ethiopia, Rwanda) in a so-called ``African-Islamic'' category and groups of countries that have little in common in one category (e.g., Kyrgyzstan and Tunisia, Bolivia and the Philippines, South Africa and the UAE), leading to misrepresentations and enforcing stereotypes.

\subsection{Set Fair and Realistic Expectations}
As pointed out by \citet{dogruoz-sitaram-2022-language}, tools for low-resource languages are often perceived as scaled-down versions of those developed for high-resource languages. Building on previous work that explores what this may mean for speakers \cite{bird-2022-local,markl-etal-2024-language}, we shift the focus to researchers and practitioners, who are often expected to build similar models to those created for high-resource languages---i.e., tackling the same NLP tasks and achieving similarly high performance.
However, this expectation can be unrealistic for several reasons, including users’ actual needs \cite{blaschke2024dialect}, the specific linguistic features of the language \cite{bird2024centering}, and the lack of funding often tied to the perceived ``prestige'' of a language, as reported by our respondents and similarly discussed by \citet{mihalcea2024ai} in the context of LLM research.

\paragraph{No Prescription}
\citet{joshi-etal-2020-state} conducted a survey on the state of NLP across various languages and found that people do not necessarily want the tools that researchers assume they need. One should not prescribe what NLP research on mid- to low-resource languages should be about. The real challenge lies in striking a balance between the technologies that local communities require and the directions pursued by the research community. This balance can be strengthened through ongoing communication and collaboration with various stakeholders \cite{lent-etal-2022-creole,mager2023ethical,bird2024centering,cooper2024s}.

\paragraph{Dealing with a ``Solved'' Problem in a New Language is an Actual Contribution}
Just because automatic systems obtain high scores on some English datasets, often making use of massive amounts of English data, does not mean that the problem is solved generally, or that working on that problem in other languages is no longer interesting or valuable.
Working with each new language presents many new challenges, 
e.g., a rich morphology or the presence of tones \cite{adebara-abdul-mageed-2022-towards}.
Such work should not be undervalued by calling it  ``a replication'' study.

\subsection{Check the Source Even if the Language is Low-resource}
Due to the limited availability of online data for some languages, there is a tendency to use any accessible source to build resources—often without considering the ethical implications or the appropriateness of the content. While it is typically more convenient to rely on religious texts, song lyrics, or film subtitles, such sources should be critically assessed \cite{hutchinson-2024-modeling, mager2023ethical}. For example, song lyrics are not representative of everyday communication \cite{mayer2008rhyme}, as they often rhyme and may contain profanity that is not typical of daily language use.

Moreover, scraping Indigenous texts without obtaining proper consent or permissions, and without sensitivity to cultural or religious contexts \cite{hutchinson-2024-modeling}, often results in work on low-resource languages being framed in utilitarian terms, with insufficient attention to deontological concerns. For instance, the use of religious texts without acknowledging their potential implications can lead to cultural misrepresentations, such as portraying certain communities as uniformly religious \cite{mager2023ethical}.

Additionally, some researchers resort to synthetic data generated via machine translation or large language models (LLMs), despite well-documented limitations, particularly in multicultural or multilingual contexts \cite{hershcovich-etal-2022-challenges} (see Section~\ref{subsubsec:marked_limitations}). Insights from other disciplines, including closely related fields such as linguistics \cite{turner2023decolonisation}, can help guide the selection of more culturally appropriate data sources, especially considering that such datasets are likely to persist and influence future research.


\subsection{Position Your Contribution}

Similar to \citet{hutchinson-2024-modeling}, we encourage the inclusion of positionality statements. Specifically, authors should indicate their relationship to the language(s) they work on. This may include, for example, their level of fluency, whether they have formally studied the language(s) in question, and whether they collaborated with native or fluent speakers for tasks such as data annotation. It is also important to clearly state the cultural background of data creators and annotators to avoid false generalisations and regional marginalisation (e.g., treating all Arabic-speaking countries as one homogeneous group \cite{keleg-2025-llm}).

\section{Conclusion}
We present insights from NLP researchers and practitioners working on under-served languages. We discuss reported limitations in this area of research and highlight issues related to data standards, along with common practices and concerning trends, such as problematic incentivisation and the lack of recognition for data workers' labor. 
\newline
We then offer actionable recommendations to improve data quality by reflecting on data sources, setting realistic expectations for under-resourced languages, and centering speakers and end users while ensuring fairness to data workers.


\section*{Limitations}
We acknowledge the the risk of selection bias.
Nonetheless, our main goal was to give voice to the concerns of data annotators and researchers working on mid- to low-resource languages.

\section*{Ethical Considerations}
While most respondents shared their contact information, it was mainly for following up on the resulting study. 
\newline
We do not share any information that may reveal their identities or the projects they reported on.\\[4pt]
{\bf Positionality Statement}\\
The three authors are affiliated with governmental and academic institutions in the UK, Sweden, and Canada, respectively. They have experience working with large, diverse groups and in developing new NLP artefacts for high-, mid-, and low-resource languages.
\newline
Nedjma Ousidhoum is linguistically proficient in Arabic (Algerian, MSA, and Classical), French, and English. She has lived in Algeria, Hong Kong, and the UK.
\newline
Meriem Beloucif is linguistically proficient in Arabic (Algerian and MSA), French, and English, and has a moderate knowledge of German and Swedish. She has lived in Algeria, Hong Kong, Denmark, Germany, and Sweden.
\newline
Saif M. Mohammad is proficient in English, Urdu, and Hindi. He has lived in India, USA, and Canada.

\section{Acknowledgments}
We thank our survey respondents for their insightful comments, and we are grateful to everyone who helped us spread the word on social media or via email.
\newline
Many thanks to Joanne Boisson, Michael Schlichtkrull, Sowmya Vajjala, and Evani Radiya-Dixit for their helpful feedback, as well as to the anonymous Action Editor and reviewers for their comments and suggestions.

\bibliography{anthology,custom}
\bibliographystyle{acl_natbib}
\newpage

\appendix

\newpage
\section*{Appendix}
\section{Questionnaire}
We would like to investigate the common practices in NLP research on low-resource languages (language variants and ``dialects'' included). 

If you are/were involved in NLP research on low-resource languages, we would like to hear from you. Note that we \textbf{**will not**} share your name or demographic information in public. We will only be checking your name for potential follow-up. 

(You can also include your initials if you do not want to disclose your name.)

\begin{itemize}
    \item Email.
    \item Name.
    \item \textit{(Optional)} Occupation/Affiliation (if any).
    \item Which languages do you work on?  Language variants and "dialects" included. Please use commas to separate the languages. E.g., language 1, language 2, ...
    \item What kind of NLP tasks are you interested in? You can name more than one.
    \item What kind of NLP tools would be relevant/useful for your language(s)?
    \item Why do you work on this/these language(s) ? You can choose more than one option.
    \begin{itemize}
        \item I have a genuine interest in languages.
        \item I want to build technologies for as many languages as possible.
        \item I want to build technologies for my language.
        \item Existing technologies in my language of interest suffered from marked limitations.
        \item I want to contribute to research on LLMs.
        \item I have a genuine interest in NLP/CL/ML.
        \item Other. [Note that this is a free text field]
    \end{itemize}
    \item \textit{(Optional)} If your answer to the previous question included "Existing technologies in my language of interest suffered from marked limitations.", can you tell us why? You can choose more than one option.
    \begin{itemize}
        \item Resources are scarce.
        \item The data is not representative of the language usage.
        \item The annotation is not performed by fluent speakers.
        \item The tools do not perform well.
        \item The tools are not aligned with the needs of the language speakers.
        \item The tools are not that useful.
        \item Other. [Note that this is a free text field]
    \end{itemize}
    \item \textit{(Optional)} If you answered "Existing technologies my language of interest suffered from marked limitations.", can you give an example of these resources or tools?
    \item \textit{(Optional)} If you answered "Existing technologies my language of interest suffered from marked limitations.", can you share why?
    \item If you were involved in previous projects, what kind of work were you involved in? 
    \begin{itemize}
        \item Annotation.
        \item Data collection.
        \item Data creation (e.g., coming up with instructions, questions, etc)
         \item Other. [Note that this is a free text field]
    \end{itemize}
    \item If you were involved in previous projects, did you often get credit for it?
    \begin{itemize}
        \item Always.
        \item Often.
        \item Sometimes.
        \item Rarely.
        \item Never.
        \item Other. [Note that this is a free text field]
    \end{itemize}
    \item \textit{(Optional)} If you were involved in the data collection and/or data annotation in previous projects, how often were you offered authorship?
    \begin{itemize}
        \item Always.
        \item Often.
        \item Sometimes.
        \item Rarely.
        \item Never.
        \item Other. [Note that this is a free text field]
    \end{itemize}
    \item \textit{(Optional)} In projects for which you did not receive financial compensation or authorship, and where you were involved in the data collection and/or data annotation, what was your incentive?
    \begin{itemize}
        \item I was part of a community.
        \item I had access to additional resources (e.g., GPUs, data, etc.).
        \item I was acknowledged on the project website.
        \item I was acknowledged in the paper.
        \item Other. [Note that this is a free text field]
    \end{itemize}
    \item \textit{(Optional)} For projects where you were simply acknowledged for being an annotator, how long did the data annotation process take?
    \begin{itemize}
        \item <=2 hours.
        \item 2-6 hours.
        \item A day of work.
        \item 1-7 days.
        \item Other. [Note that this is a free text field]        
    \end{itemize}
    \item Are you part of a community? (Yes/No)
    \item \textit{(Optional)} If you are part of a community, can you name it?
    \item \textit{(Optional)} Were you involved in projects with industry or academia?
    \begin{itemize}
        \item Industry.
        \item Academia.
        \item Both.
    \end{itemize}
    \item \textit{(Optional long text answer)} Can you name the institutions/projects? (We will not make the names public if you do not want to share them publicly. See question below.)
    \item Are you happy making the project names public? (Yes/No)
    \item \textit{(Optional long text answer)} What are the potential challenges that the NLP/CL community working on the languages that you mentioned face?
    \item Would you like to receive updates about this project? (Yes/No)
\end{itemize}

\section{Languages}
The full list of the languages that our respondents have worked is included in the following. Note that participants could work on more than one language. Some may have also conducted research for high-resource languages and the respondents may or may not speak the language(s).
\paragraph{Named Mid- to Low-resource Languages}
Afaan Oromo, Albanian, Algerian Arabic, Amharic, Assamese, Awigna, Azerbaijani, Bangla, Basque, Bikol, Cebuano, Coptic, Creole, Croatian, Danish, Egyptian Arabic, Emakhuwa, Faroese, Filipino, Geez, Greek, Harari, Hausa, Hindi, Igbo, Ilocano, Indonesian, Irish, IsiXhosa, Kanuri, Kazakh, Kinyarwanda, Kiswahili, Korean, Light Warlpiri, Lingala, Luganda, Luhya (Lumarachi dialect), Malaysian English, Marathi, Moroccan Arabic, Nepalese, Nyanja, Oromo, Persian/Farsi, Pidgin, Punjabi, Raramuri
Russian, Saudi Arabic, Sena, Setswana, Sundanese, Swahili, Tagalog, Tarifit Berber, Tigrinya, Tsonga, Tunisian Arabic, Turkish, Urdu, Warlpiri, Welsh, Wixarika, Wolof, Xhosa, Yoreme Nokki, Yorùbá, Zulu.
\paragraph{Families of Languages}
African languages, Arabic dialects/variations, English variants, Chatino languages, Gaeilge (including all dialects), Latin American Spanish, Indian languages, Indonesian languages, Nahuatl languages, North African dialects, South East Asian languages.
\paragraph{Named High-resource Languages}
English, French, Italian, Modern Standard Arabic, Spanish.

\end{document}